\documentclass{article}
\usepackage{amsmath,graphicx,mlspconf,booktabs}
\usepackage{xcolor}
\usepackage{multirow}
\usepackage{amssymb}
\usepackage[colorlinks]{hyperref}
\usepackage[caption=false,font=footnotesize]{subfig}

\toappear{2025 IEEE International Workshop on Machine Learning for Signal Processing, Aug.\ 31-- Sep.\ 3, 2025, Istanbul, Turkey}

\title{Enhanced Arabic Text Retrieval with Attentive Relevance Scoring}

\name{%
    \begin{tabular}{c}
         Salah Eddine Bekhouche$^{1}$ \qquad Azeddine Benlamoudi$^{2}$ \qquad Yazid Bounab$^{3}$ \\
         Fadi Dornaika$^{1,4}$ \qquad Abdenour Hadid$^{3,5}$ \thanks{The supports of TotalEnergies and Sorbonne University Abu Dhabi are fully acknowledged.}
    \end{tabular}
}

\address{%
$^{1}$ University of the Basque Country UPV/EHU, San Sebastian, Spain \\%
$^{2}$ Lab. de Genie Electrique (LAGE), University of Ouargla, Ouargla, Algeria \\%
$^{3}$ Faculty of Pharmacy, Helsinki University, Helsinki, Finland \\%
$^{4}$ IKERBASQUE, Basque Foundation for Science, Bilbao, Spain\\%
$^{5}$ Sorbonne University Abu Dhabi, Abu Dhabi, UAE%
}

\begin{document}

\maketitle

\begin{abstract}
Arabic poses a particular challenge for natural language processing (NLP) and information retrieval (IR) due to its complex morphology, optional diacritics and the coexistence of Modern Standard Arabic (MSA) and various dialects. Despite the growing global significance of Arabic, it is still underrepresented in NLP research and benchmark resources. In this paper, we present an enhanced Dense Passage Retrieval (DPR) framework developed specifically for Arabic. At the core of our approach is a novel Attentive Relevance Scoring (ARS) that replaces standard interaction mechanisms with an adaptive scoring function that more effectively models the semantic relevance between questions and passages. Our method integrates pre-trained Arabic language models and architectural refinements to improve retrieval performance and significantly increase ranking accuracy when answering Arabic questions. The code is made publicly available at \href{https://github.com/Bekhouche/APR}{GitHub}.
\end{abstract}

\section{Introduction}
\label{sec:introduction}
\vspace{-1mm}
Arabic, one of the most widely spoken languages globally, presents unique linguistic challenges for natural language processing (NLP) and information retrieval (IR). Applying Dense Passage Retrieval (DPR) \cite{karpukhin2020dpr} to Arabic opens new avenues but also introduces unique challenges. Its rich morphology, frequent use of diacritics, syntactic complexity, and the coexistence of Modern Standard Arabic (MSA) with diverse dialects pose significant difficulties for conventional retrieval systems, which often struggle with normalization and semantic understanding. Despite its global importance, Arabic remains underrepresented in NLP research, resulting in fewer dedicated resources and tools compared to other major languages.

While the advent of pre-trained transformer models like AraBERT, MARBERT, and multilingual BERT has significantly advanced Arabic NLP, creating highly effective DPR systems for Arabic requires more than just fine-tuning these general models on downstream tasks. Recognizing these limitations and the lack of dedicated Arabic DPR models, recent research, exemplified by the development of AraDPR~\cite{abdallah2024arabicaqa}, has focused on training specialized retrievers. AraDPR utilizes AraBERT as its backbone and employs a contrastive learning approach, trained specifically on a combination of translated benchmark datasets and native Arabic question-answering datasets. This method has produced the first publicly available, contrastively trained Arabic DPR model, demonstrating state-of-the-art performance on Arabic passage retrieval benchmarks and significantly outperforming both traditional methods and fine-tuned multilingual or cross-lingual models in Arabic open-domain QA contexts~\cite{abdallah2024arabicaqa}. This was a crucial step towards building more robust and linguistically nuanced dense retrieval systems tailored specifically for the Arabic language.

However, even state-of-the-art dense retrieval models often rely on simple vector similarity measures (e.g., dot product or cosine similarity) for the final relevance scoring between query and passage representations. These standard scoring functions may not fully capture the intricate semantic relationships and account for the morphological variations inherent in Arabic text. To address this limitation and enhance semantic matching capabilities, we present a novel system incorporating Attentive Relevance Scoring (ARS). ARS replaces the typical final scoring step based on simple vector similarity with a more adaptive and semantically aware scoring function designed to better model relevance in the context of Arabic's linguistic features.

The potential applications for effective Arabic DPR are vast, including enhanced question-answering systems, digital libraries, Arabic-language search engines, and conversational agents for Arabic-speaking users. With the continuous increase of annotated Arabic datasets and targeted research efforts, the effectiveness and usability of Arabic DPR systems are expected to improve significantly. The main contributions of this work are:
\begin{itemize}
    \item We propose an enhanced dense retrieval framework for Arabic, integrating lightweight transformer models with an adaptive scoring mechanism.
    \item We introduce an ARS module to improve semantic matching beyond conventional vector similarity functions used in standard DPR.
    \item We present extensive experimental analysis, pointing out some limitations and future directions.
\end{itemize}

This rest of this paper is structured as follows: Section~\ref{sec:related_work} overviews the foundational work on DPR in general and Arabic DPR in particular. In Section~\ref{sec:proposed_approach}, we describe our methodology, highlighting the architectural advantages of the ARS. Section~\ref{sec:experiments} presents the experimental validation of our approach, providing detailed results and comparison against some existing works. It also offers an in-depth analysis of the findings, discussing their implications, limitations, and potential future research directions. Finally, Section~\ref{sec:conclusion} summarizes our key findings and contributions to the field.

\vspace{-3mm}
\section{Related Work}
\label{sec:related_work}
\vspace{-1mm}

The field of IR has been significantly transformed by advancements in deep learning, particularly the emergence of pre-trained transformer models \cite{devlin2019bert, vaswani2017attention}. A key development stemming from this is DPR, which has become a cornerstone for modern open-domain Question Answering (QA) and search systems. 
The core idea of DPR, popularized by \cite{karpukhin2020dpr}, involves encoding both user queries and text passages into low-dimensional, dense vector representations using dual-encoder architectures, typically based on BERT or its variants. Relevance scoring is then performed by calculating the similarity (e.g., dot product or cosine similarity) between the query vector and passage vectors. Retrieval is efficiently handled using Approximate Nearest Neighbor (ANN) search algorithms over the pre-computed passage embeddings \cite{johnson2019billion}. This paradigm demonstrated substantial improvements over traditional sparse retrieval methods like BM25 \cite{robertson2009probabilistic} on various benchmarks. Subsequent research focused on refining DPR, particularly through improved training strategies like sophisticated negative sampling techniques (e.g., ANCE \cite{xiong2020ance}, RocketQA \cite{qu2021rocketqa}) that dynamically mine hard negatives to create more robust models.
While effective, standard DPR models based on large transformers can be computationally intensive. This has spurred research into more efficient architectures. ColBERT \cite{khattab2020colbert} introduced a late-interaction mechanism, calculating relevance based on fine-grained token-level interactions between query and passage embeddings. This allows for richer semantic matching while maintaining efficiency through pre-computation and optimized vector similarity search \cite{santhanam2021colbertv2}. Other approaches like SPLADE \cite{formal2021splade} explore learned sparse representations, bridging the gap between traditional lexical matching and dense semantic retrieval, often offering competitive performance with better efficiency. Furthermore, model compression techniques like knowledge distillation (e.g., DistilBERT \cite{sanh2019distilbert}, TinyBERT \cite{jiao2019tinybert}) are increasingly explored to create lightweight yet powerful retrieval models suitable for resource-constrained environments. 

Applying these advanced retrieval techniques to Arabic poses unique challenges due to its complex morphology (rich inflection and derivation), orthographic variations (e.g., optional diacritics, inconsistent spelling of certain letters), and dialectal diversity \cite{ezzeldin2012survey}. Significant progress in Arabic NLP has been driven by the development of dedicated Arabic pre-trained language models, including AraBERT \cite{antoun2020arabert}, MARBERT \cite{abdul2020arbert}, ARBERT \cite{abdul2020arbert}, AraELECTRA \cite{antoun2020araelectra}, and CamelBERT \cite{inoue2021interplay}, which are trained on large Arabic corpora and better capture the language's nuances compared to multilingual models. However, developing effective dense retrieval systems specifically for Arabic has lagged behind English. Traditional Arabic IR systems often rely heavily on morphological analysis and query expansion techniques to handle lexical gaps \cite{darwish2014arabic}. The direct application of standard DPR techniques, even when fine-tuning multilingual models, often yields suboptimal results due to the language's specific characteristics \cite{abdallah2024arabicaqa}. Recognizing this gap, recent work has focused on creating dedicated Arabic dense retrieval models. A landmark contribution is AraDPR \cite{abdallah2024arabicaqa}, the first publicly available, contrastively trained DPR model specifically for Arabic. Utilizing AraBERT as its backbone and trained on a mix of translated benchmarks and native Arabic QA datasets (like ARCD \cite{obeidat2024arquad}), AraDPR established a new state-of-the-art for Arabic passage retrieval, significantly outperforming both traditional methods and general-purpose fine-tuned models. This highlights the importance of language-specific training and architectures for achieving high performance in Arabic retrieval contexts, also emphasized by cross-lingual benchmark efforts like TyDi QA \cite{clark2020tydi}. 
Despite the success of models like AraDPR, standard dense retrieval predominantly relies on a single vector representation for queries and passages, with relevance determined by simple geometric similarity (dot product/cosine). While computationally efficient, this approach might oversimplify the complex relevance judgments required, especially for morphologically rich and semantically nuanced languages like Arabic. While late-interaction models like ColBERT \cite{khattab2020colbert} address this by allowing token-level interactions, other neural ranking frameworks have long explored learned interaction layers on top of global embeddings to improve upon simple similarity measures \cite{huang2013learning, gupta2019faq}. Our work builds on this latter paradigm, proposing a lightweight, adaptive scoring module specifically tailored for the challenges of Arabic retrieval.

\vspace{-3mm}
\section{Proposed Approach}
\label{sec:proposed_approach}
\vspace{-1mm}

\begin{figure}
    \centering
    \includegraphics[width=0.9\linewidth]{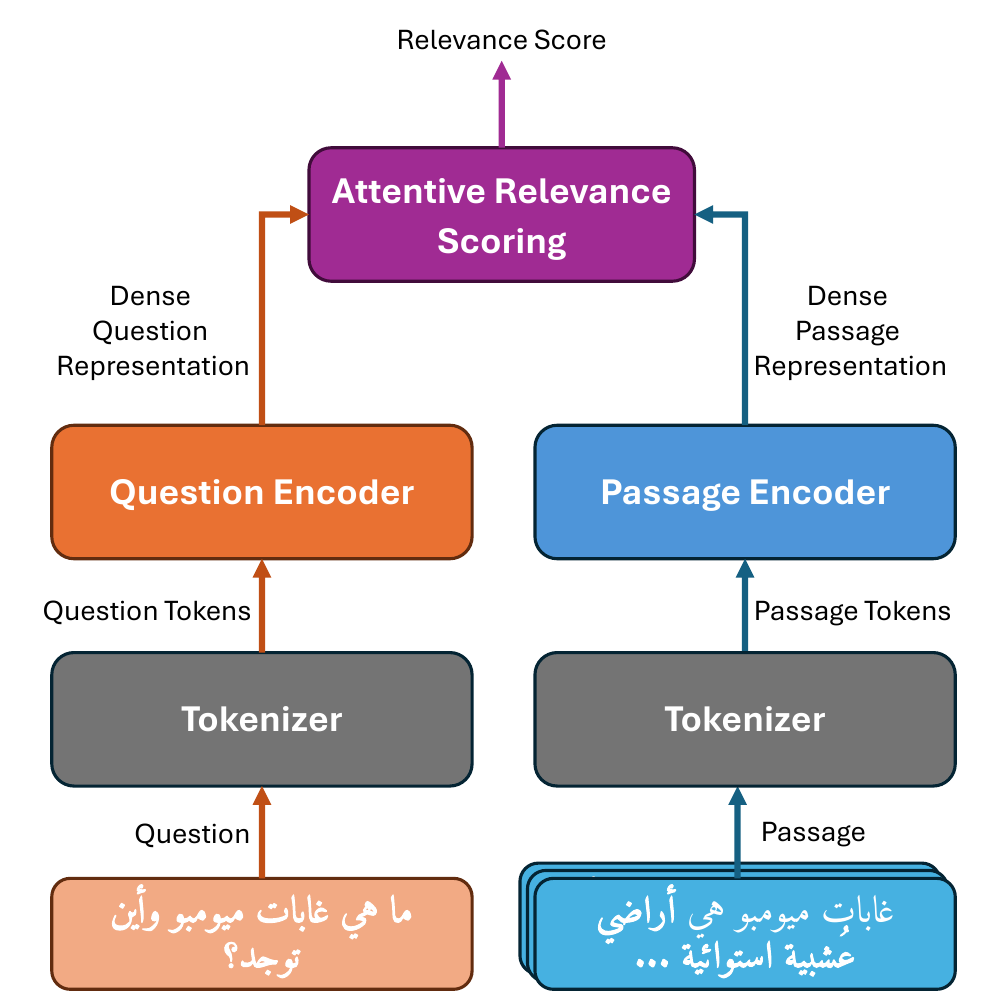}
    \caption{Overview of the proposed APR framework.}
    \label{fig:general_architecture}
\end{figure}

This section introduces our proposed method, called Adaptive Passage Retrieval (APR), tailored for Arabic text retrieval. Our approach builds upon the DPR framework \cite{karpukhin2020dense} and incorporates a lightweight, Arabic-specific encoder (MiniBERT)~\cite{safaya2020kuisail} alongside a novel scoring mechanism termed Attentive Relevance Scoring (ARS). This combination is designed to enhance retrieval accuracy by modeling enhanced semantic interactions in Arabic texts, while maintaining computational efficiency suitable for low-resource settings. The overall architecture of our pro APR is illustrated in Figure~\ref{fig:general_architecture}.

\subsection{Dual-Encoder Architecture}

APR employs a dual-encoder architecture consisting of a \textit{question encoder} ($E_Q$) and a \textit{passage encoder} ($E_P$), both initialized with weights from MiniBERT, a transformer model pre-trained on Arabic corpora. This initialization allows the system to better capture Arabic linguistic features. Given a question $q$ and a passage $p$, each encoder processes its respective input independently, generating two types of representations: sequence-level and pooled [CLS] token representations.

The sequence-level representations, denoted $\mathbf{H}_q \in \mathbb{R}^{B \times L \times d}$ and $\mathbf{H}_p \in \mathbb{R}^{B \times L \times d}$, capture contextual embeddings for each token in the question and passage, respectively. Here, $B$ is the batch size, $L$ is the input lengths, and $d$ is the hidden dimension of the MiniBERT model.

For global semantic representation, we extract the final-layer hidden state corresponding to the special [CLS] token, and apply $\ell_2$ normalization to obtain fixed-size embeddings:
\begin{equation}
\label{eq:cls_embeddings}
    \mathbf{q} = \text{Norm}(E_Q(q)_{[\text{CLS}]}) \in \mathbb{R}^d, \quad
    \mathbf{p} = \text{Norm}(E_P(p)_{[\text{CLS}]}) \in \mathbb{R}^d,
\end{equation}
where $\text{Norm}(\cdot)$ denotes $\ell_2$ normalization. This normalization projects vectors onto a unit hypersphere, which is beneficial for stabilizing contrastive similarity computations.

\subsection{Attentive Relevance Scoring}

The ARS module computes an adaptive semantic similarity between query and passage embeddings through a trainable interaction model.

First, the embeddings are projected into a shared space using:
\begin{equation}
    \mathbf{h}_q = W_q \mathbf{q}, \quad
    \mathbf{h}_p = W_p \mathbf{p},
\end{equation}
where $W_q, W_p \in \mathbb{R}^{h \times d}$ are learnable projection matrices and $h$ is the shared hidden dimensionality.

Next, we model interactions using element-wise multiplication ($\odot$) followed by a non-linear activation to compute the interaction vector $\mathbf{a}$:

\begin{equation}
\label{eq:interaction_vector_a}
    \mathbf{a} = \tanh(\mathbf{h}_q \odot \mathbf{h}_p).
\end{equation}
Here, $\tanh(\cdot)$ is the hyperbolic tangent function, which applies a non-linearity to the element-wise product of the projected embeddings.

Finally, a scalar relevance score $r$ is computed via an attention vector $w_a \in \mathbb{R}^h$:
\begin{equation}
r = \sigma\left( w_a^\top \mathbf{a} \right),
\end{equation}
where \( \sigma(\cdot) \) is the sigmoid function.

During inference, the passage embeddings $\mathbf{p}$ (as defined in Equation~\ref{eq:cls_embeddings}) are pre-computed for all $N_P$ documents in the knowledge source to enable efficient query processing. When a new query $\mathbf{q}$ arrives, its embedding is generated in real time. The ARS module then computes a relevance score $r^{(j)}$ by interacting the query embedding with each of the pre-computed passage embeddings ${\mathbf{p}^{(j)}}{j=1, ..., N_P}$. Finally, all $N_P$ passages are ranked based on the resulting ARS scores ${r^{(j)}} $ to produce the retrieval output.

\subsection{Loss Function}
To optimize both overall and fine-grained semantic alignment, we define a total loss function:

\begin{equation}
\mathcal{L}_{\text{total}} = \alpha \cdot \mathcal{L}_{\text{cons}} + \beta \cdot \mathcal{L}_{\text{dyn}} + \gamma \cdot \mathcal{L}_{\text{reg}},
\end{equation}

where $\alpha = 1$, $\beta = 1$, and $\gamma = 0.1$ are empirically determined weights. During training, each batch contains $B$ queries, with each query paired with one positive and a pool of 29 negative passages, ensuring diverse negative exposure while maintaining efficiency.

\subsubsection{Contrastive Loss ($\mathcal{L}_{\text{cons}}$)}

We use a contrastive loss based on InfoNCE. It works on the \texttt{[CLS]} token embeddings. This loss helps the model align the query embedding ($\mathbf{q}$) with the correct passage embedding ($\mathbf{p}^+$) and separate it from the incorrect ones ($\mathbf{p}^-$):

\begin{equation}
    \mathcal{L}_{\text{cons}} = -\frac{1}{B} \sum_{i=1}^{B} \log\left( \frac{ \exp(\mathbf{q}_i^\top \mathbf{p}_i^+ / \tau) }{ \exp(\mathbf{q}_i^\top \mathbf{p}_i^+ / \tau) + \sum_{j=1}^{N} \exp(\mathbf{q}_i^\top \mathbf{p}_{i,j}^- / \tau) } \right),
\end{equation}

Here, $B$ is the batch size. $\mathbf{q}_i$ is the embedding for the $i$-th query. $\mathbf{p}_i^+$ is the embedding for the correct (positive) passage. $\{\mathbf{p}_{i,j}^-\}_{j=1}^N$ are the embeddings for $N$ negative passages (N=29). The temperature parameter $\tau > 0$ is a learnable parameter used to scale the dot products.

\subsubsection{Dynamic Relevance Loss ($\mathcal{L}_{\text{dyn}}$)}

Standard contrastive learning separates positive and negative pairs, but it might not fully capture small differences in the scores. This is especially important in Arabic, where similar words can have slightly different meanings. To handle this, we add a dynamic relevance loss that supervises the model’s relevance scores (ARS scores, $r$).

This loss aims to increase the score $r_i^+$ for the correct passage and decrease the score $r_i^-$ for the incorrect one. At the same time, it encourages a wider range of scores across the batch. This prevents the scores from becoming too similar, which can happen with difficult negative passages.

The formulation for $\mathcal{L}_{\text{dyn}}$, averaged over the batch of size $B$, is:

\begin{align}
    \mathcal{L}_{\text{dyn}} =\ & -\frac{1}{B} \sum_{i=1}^{B} \left[ \log(r_i^+ + \epsilon) + \log(1 - r_i^- + \epsilon) \right]
\end{align}

Here, $r_i^+$ and $r_i^-$ are the ARS scores (typically passed through a sigmoid, and thus constrained between 0 and 1) for the positive and negative passages, respectively. $\epsilon$ is a small constant (e.g., $10^{-8}$) added for numerical stability. The negative passage is randomly selected from the pool of negative passages. This loss aims to maximize $r_i^+$ and minimize $r_i^-$, ideally pushing $r_i^+ \rightarrow 1$ and $r_i^- \rightarrow 0$ for each sample in the batch. This encourages the model to produce confident and well-separated scores, improving the quality of passage retrieval.

\subsubsection{Relevance Score Logit Regularization ($\mathcal{L}_{\text{reg}}$)}

As an additional step, we apply a regularization loss on the raw relevance scores before applying the sigmoid function (called logits, $s$). This helps keep the training stable and prevents all outputs from becoming too similar. 

While $\mathcal{L}_{\text{dyn}}$ encourages variance in the final ARS scores ($r^+$ and $r^-$), $\mathcal{L}_{\text{reg}}$ does the same for the raw logits ($s^+$ and $s^-$). This helps maintain a useful range of values before activation, supports gradient flow, and prevents the model from becoming too confident or producing uniform predictions, especially early in training or with hard negatives.

The regularization loss is:
\begin{equation}
    \mathcal{L}_{\text{reg}} =\operatorname{Std}(s_{\text{batch}}^+) + \operatorname{Std}(s_{\text{batch}}^-),
\end{equation}

Here, $s_{\text{batch}}^+$ and $s_{\text{batch}}^-$ are the sets of raw, pre-sigmoid relevance scores (logits) for positive and negative examples in the batch. $\operatorname{Std}(\cdot)$ means standard deviation. 

\section{Experiments}
\label{sec:experiments}

\subsection{Dataset}

For our experiments, we used the ArabicaQA dataset, a comprehensive, human-annotated Arabic question answering corpus specifically designed for open-domain retrieval and machine reading comprehension tasks. The dataset\footnote{\href{https://huggingface.co/datasets/abdoelsayed/Open-ArabicaQA/tree/main/retreiver/human-annotated}{Open-ArabicaQA Human-Annotated Retriever Dataset}} is divided into standard training, validation and test subsets. The training set consists of 58,727 question-answer pairs, each accompanied by a relevant positive passage with the answer and 29 hard negative passages. These hard negative passages are semantically similar to the question, but do not contain the correct answer, which makes them valuable for training robust retrieval models. The validation set contains 12,722 question-answer pairs and the test set contains 12,597 question-answer.

The textual knowledge source for ArabicaQA is derived from the Arabic Wikipedia\footnote{\href{https://huggingface.co/datasets/abdoelsayed/Open-ArabicaQA/tree/main/wikipedia_split}{Open-ArabicaQA Wikipedia Split Dataset}}, which contains approximately 1,222,923 articles. These articles serve as the source passages from which answers are retrieved or extracted. During performance evaluation on both the validation and test sets, each question is evaluated against the full Wikipedia split, ensuring a realistic and comprehensive assessment of retrieval and reading comprehension capabilities.

\vspace{-3mm}
\subsection{Experimental Setup}
All experiments were conducted on a machine with six NVIDIA L4 GPUs, each providing 24 GB of VRAM. A multi-GPU distributed training strategy was employed to accelerate the training process. Mixed precision training was not utilized, and the gradient accumulation step was set to 1 to maintain stable gradient updates.

The model architecture comprises a question encoder and a context (passage) encoder, each containing approximately 11.55 million parameters. An auxiliary ARS module introduces an additional 0.13 million parameters, resulting in a total of approximately 23.23 million trainable parameters for the complete APR model.

Training was performed on the training split of the ArabicaQA dataset, while model validation was conducted on the validation split, consisting of 12,722 examples. Both training and validation phases used a per-GPU batch size of 32.

Optimization was carried out using the AdamW optimizer~\cite{loshchilov2019decoupled} with a fixed learning rate of $1 \times 10^{-4}$ and an $\epsilon$ value of $1 \times 10^{-8}$. A linear learning rate scheduler was adopted, linearly increasing the learning rate from an initial factor of 0.1 to the target value over the course of training. To ensure numerical stability during training, gradient clipping was applied with a maximum norm of 1.0.

\subsection{Results and Discussion}

\begin{figure}[htb]
\centering
\includegraphics[width=1.00\linewidth]{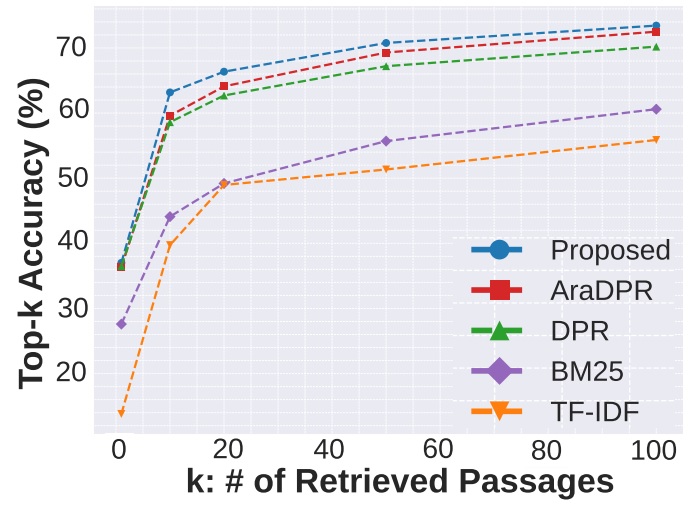}%
\caption{Top-$k$ retrieval accuracy comparison on the validation set of the ArabicaQA dataset. Our APR model demonstrates consistent improvements across all values of $k$ compared to existing methods.}
\label{fig:topk_accuracy}
\end{figure}

Figure~\ref{fig:topk_accuracy} illustrates the top-k retrieval accuracy on the validation set, while Table~\ref{tab:retriever_acc} reports the performance on the test set, comparing our results with state-of-the-art methods on the ArabicaQA dataset.

\begin{table}[btb]
\small
    \caption{Retriever Module Performance Comparison (Top-$k$ accuracy on test set)}
    \label{tab:retriever_acc}
    \centering
    \begin{tabular}{lccccc}
    \toprule
    \textbf{Method} & \textbf{Top-1} & \textbf{Top-10} & \textbf{Top-20} & \textbf{Top-50} & \textbf{Top-100} \\
    \midrule
    TF-IDF \cite{mozannar2019neural} & 14.35 & 40.86 & 46.87 & 51.71 & 55.36 \\
    BM25 \cite{robertson2009probabilistic} & 28.70 & 43.40 & 48.20 & 54.60 & 59.30 \\
    DPR \cite{karpukhin2020dpr} & 36.40 & 57.80 & 62.10 & 66.60 & 69.50 \\
    AraDPR \cite{abdallah2024arabicaqa} & 36.10 & 58.40 & 63.40 & 68.60 & 71.90 \\
    \textbf{APR (Ours)} & \textbf{37.01} & \textbf{63.17} & \textbf{66.36} & \textbf{70.77} & \textbf{73.43} \\
    \bottomrule
    \end{tabular}
\end{table}
    
As shown in Table~\ref{tab:retriever_acc}, our proposed APR model outperforms all baseline systems across all top-$k$ retrieval thresholds ($k = 1, 10, 20, 50, 100$). Compared to AraDPR—the strongest Arabic baseline—APR achieves absolute gains of +0.91\% in Top-1, +4.77\% in Top-10, and +1.53\% in Top-100 accuracy. These consistent improvements demonstrate that APR effectively leverages the ARS module to better distinguish truly relevant passages from semantically similar but incorrect ones.
    
Furthermore, Figure~\ref{fig:topk_accuracy} visually emphasizes the strong performance of APR at increasing values of $k$. While the performance gap between APR and the other retrievers is modest at lower $k$ values, it gradually increases as $k$ increases, indicating superior ranking capabilities and a better understanding of the relevance of answers within the large document collection. More specifically, the figure shows that APR achieves an accuracy of about 38\% at $k=5$, which is slightly ahead of dense baseline methods such as AraDPR (36.10\%) and DPR (36.40\%), and significantly higher than sparse methods such as BM25 (28.70\%) and TF-IDF (14.35\%). As $k$ increases, all dense models — including APR, AraDPR and DPR — begin to plateau at $k=50$, suggesting that most relevant passages are found early in the ranking process. Despite this saturation, APR maintains a consistent lead over AraDPR and DPR at each $k$ value, highlighting the benefits of response-based ranking. Ultimately, APR achieves the highest top-$k$ accuracy of 75.01\% at $k=100$, demonstrating its robustness even for broader retrieval scopes.

These results highlight the strength of incorporating Answer Relevance Scoring into dense retrieval. The higher Top-$k$ accuracy, particularly at low cutoffs such as Top-1 and Top-10, ensures that downstream reader modules receive higher-quality candidates, thereby improving the overall question answering pipeline.

\vspace{-2mm}
\section{Conclusion}
\label{sec:conclusion}
\vspace{-2mm}
We presented an improved dense retrieval framework tailored for Arabic question answering. By integrating a lightweight encoder and ARS, our approach addresses key challenges in Arabic IR, including morphological complexity and limited semantic generalization in traditional models. This work opens new directions for efficient and accurate retrieval in underrepresented languages and lays the foundation for future enhancements in Arabic-language AI systems. While our results are promising, we recognize that we need a more detailed analysis. Future work will focus on doing thorough ablation studies to separate the contributions of our proposed components and conducting a qualitative analysis to provide deeper insights into the model's practical strengths.

\vspace{-2mm}
\small
\bibliographystyle{IEEEbib}
\bibliography{references}

\end{document}